\documentclass[sigconf,nonacm]{acmart}

\AtBeginDocument{%
  }

\usepackage{amsmath}
\usepackage{booktabs}
\usepackage{tabularx}
\usepackage{array}
\usepackage{placeins}
\setcopyright{none}
\renewcommand\footnotetextcopyrightpermission[1]{}

\begin{document}

\title[AutoLR]{AutoLR: Automating the Path from Research to Launch Review in Industrial Recommender Systems}

\author{Qi Zhang}
\affiliation{%
  \institution{NetEase, Inc.}
  \country{China}}
\email{zhangqi21@corp.netease.com}

\author{Yanlin Chen}
\affiliation{%
  \institution{NetEase, Inc.}
  \country{China}}
\email{alex01@corp.netease.com}

\author{Wenchao Xiao}
\affiliation{%
  \institution{NetEase, Inc.}
  \country{China}}
\email{xiaowenchao@corp.netease.com}

\renewcommand{\shortauthors}{Zhang et al.}

\begin{abstract}
Improving an industrial recommender is an iterative research-and-engineering process rather than a direct path from idea to deployment. In \textbf{DASHEN, NetEase's gaming-community app}, algorithm engineers typically identify promising directions from research papers, technical reports, and prior production experiments; reproduce or adapt the underlying methods; implement them in the production codebase; and evaluate the resulting models through training and offline experiments. Promising candidates are then advanced to online A/B tests, and those demonstrating robust gains are submitted to Launch Review---the internal gate for full-traffic rollout. Large language models (LLMs) can assist with individual stages of this workflow, but the overall process remains human-dependent without a harness that can reliably coordinate them across long-running, often multi-day experimental cycles.

We present \textbf{AutoLR}, initially built as \textbf{Auto Launch Review} and later extended upstream into an autonomous research-to-launch harness. 
AutoLR combines three system mechanisms: a \textbf{multi-expert council} that debates and adversarially reviews proposals; 
a \textbf{deterministic evidence-weighted exploration--exploitation selector} that allocates a limited trial budget across candidate directions and uses Council reranking; 
and a layered knowledge system that combines external research, production-system knowledge, and DASHEN-specific domain knowledge---such as game communities, player characteristics, and content-interaction patterns---with posterior evidence from configurations, patches, logs, failures, and offline outcomes. 
LLM agents perform semantic reasoning and code generation, while deterministic controllers retain authority over execution, metric extraction, guardrails, and persistent state transitions.

A path-aware audit of several months of experiment logs identified \textbf{1,586 unique completed evaluations} across two DASHEN recommendation scenarios: the single- and two-column feed and the immersive-video feed. Across these scenarios, nine production Launch Reviews documented positive online lifts. For descriptive reporting, the signed relative-lift values across these heterogeneous records sum arithmetically to \textbf{+5.75\% in content-consumption penetration rate}, \textbf{+10.83\% in total content-consumption time}, and \textbf{+5.55\% in total valid content views (VV)}. These sums are descriptive and are not pooled treatment effects.

Following harness- and loop-level optimizations, routine AutoLR iterations were migrated from Claude Opus-class models to a mixed DeepSeek-V4-Pro/Flash stack and ran stably at an observed LLM API cost of \textbf{RMB 3--4 per iteration}, excluding model-training compute and internal infrastructure.\footnote{The reported LLM API cost is calculated using API prices in effect before 17 August 2026.}

We also identify a reliability risk for autonomous Launch Review. Once an accepted offline measurement becomes the reference for later experiments, noisy promotions can accumulate through a \textbf{KEEP ratchet}, producing a monotone trajectory that resembles progress. AutoLR therefore shows both the promise and the central design requirement of autonomous recommendation development: search broadly, but allow persistent state changes only under calibrated, deterministic evidence.
\end{abstract}


\maketitle

\section{Introduction}
\label{sec:introduction}

Industrial recommender systems improve through repeated experimentation. Before a candidate is fully rolled out, algorithm engineers typically complete a long path: literature research, business analysis, paper reproduction, hypothesis generation, code implementation, training, offline evaluation, limited-traffic online A/B testing, and final result attribution. In DASHEN, these artifacts are assembled at \textbf{Launch Review}, where offline evidence, online effects, business guardrails, and implementation readiness are used to decide whether a candidate should receive full traffic.

The bottleneck is not any one step. It is the repeated handoff among research, code, training platforms, logs, metrics, and business decisions. As a recommendation system matures, expected gains become smaller while the number of plausible ideas grows. Compute may support more parallel trials, but the range of ideas a team can explore remains constrained by human attention, engineering time, and the ability to preserve lessons across experiments.

LLMs create an opportunity to reorganize this workflow. Recent ML-engineering systems and benchmarks demonstrate a growing ability to interpret task specifications, write or modify training code, execute experiments, and iteratively refine candidate solutions using validation results and execution feedback \citep{chan2025mlebench,jiang2025aide,nam2025mlestar}. Related scientific-discovery systems extend this loop to research-idea generation and evaluator-guided algorithmic search \citep{lu2024aiscientist,novikov2025alphaevolve}.

Recent work has begun to carry this pattern into industrial recommendation. Google's Self-Evolving Recommendation System uses an Offline Agent, or Fast Loop, to generate and screen hypotheses against proxy metrics, and an Online Agent, or Slow Loop, to validate selected candidates against delayed north-star metrics; the system reports several production launches at YouTube \citep{wang2026selfevolving}. Kuaishou's AgentX organizes a four-stage closed loop spanning proposal generation, production-code implementation, guarded online A/B evaluation, and harness evolution from accumulated execution trajectories \citep{lao2026agentx}. RecHarness decouples edit-direction selection from concrete code mutation: validation-driven Thompson sampling allocates trials across predefined edit arms, while an LLM generates executable modifications within the selected direction \citep{ling2026recharness}. Meta's engineering report on the Ranking Engineer Agent (REA) describes a planner--executor architecture for managing asynchronous ads-ranking experiments across multi-day to multi-week workflows \citep{kumar2026rea}.

These systems also reveal that the main design problem is broader than choosing a stronger foundation model. A production agent needs a \textbf{harness} that controls context, tools, experiment state, execution, recovery, memory, and authority. A capable LLM does not by itself determine which proposal deserves the next trial, whether a patch respects a living repository, when a failure should be retried, which metric is authoritative, or when an observed improvement is strong enough to change persistent state. Long-running autonomy therefore depends on explicit state and bounded, auditable interfaces \citep{rajasekaran2026harness,kumar2026rea}.

AutoLR is named after the endpoint it is designed to reach. \textbf{Auto Launch Review} refers to automating the path from a recommendation optimization request to a Launch Review-ready candidate, rather than automating the review meeting itself. Given a business objective and repository constraints, AutoLR autonomously retrieves evidence, proposes and debates hypotheses, selects directions, modifies and verifies code, trains and evaluates models, and packages promising candidates.

In the deployment studied here, AutoLR operates autonomously through offline evaluation and candidate packaging. Engineers still select candidates for online A/B testing, and Launch Review remains the human gate for full-traffic rollout. Results from these stages are written back as production evidence for subsequent research cycles. The long-term goal is to extend the same governed automation through online experimentation and rollout:

\begin{equation}
\begin{aligned}
\text{Research}
&\rightarrow \text{Debate}
\rightarrow \text{Select}
\rightarrow \text{Implement} \\
&\rightarrow \text{Verify}
\rightarrow \text{Train}
\rightarrow \text{Offline Evaluate} \\
&\rightarrow \text{Package}
\rightarrow \boxed{\text{Engineer Review}} \\
&\rightarrow \text{Online A/B}
\rightarrow \boxed{\text{Launch Review}} \\
&\rightarrow \text{Rollout}
\rightarrow \text{Learn}
\rightarrow \text{Repeat}.
\end{aligned}
\label{eq:loop}
\end{equation}

The enduring abstraction is the governed lifecycle rather than a fixed set of agents: models and tools may change, while the harness preserves evidence contracts, state transitions, guardrails, recovery policies, and authority boundaries.

This paper makes four contributions:

\begin{enumerate}

    \item \textbf{A governed, lifecycle-centered harness.}
    AutoLR structures the path from research to Launch Review and autonomously executes its upstream stages, including evidence retrieval, proposal generation, repository-grounded implementation, verification, training, offline evaluation, and candidate packaging, while preserving human authority over online A/B testing and full-traffic rollout.

    \item \textbf{An evidence-grounded proposal and trial allocation process.}
    A role-specialized multi-expert council independently generates, debates, and adversarially reviews candidate directions, while a deterministic, evidence-weighted exploration--exploitation selector allocates limited trials by combining prior experimental outcomes with council reranking.

    \item \textbf{A layered knowledge and experiment memory.}
    AutoLR grounds agents in external research, production-system facts, and DASHEN-specific domain knowledge, and accumulates structured evidence from prior configurations, code patches, training logs, failures, offline results, and production outcomes to guide subsequent iterations.

    \item \textbf{Production-scale evaluation and reliability findings.}
    A path-aware audit identifies \textbf{1,586 unique completed evaluations} across two recommendation scenarios, while nine production Launch Review records document online outcomes for engineer-selected candidates. We further report the migration of routine workloads to lower-cost models and characterize the \textbf{KEEP ratchet}, a stateful selection failure mode in which noise-driven promotions alter the experiment baseline and can create non-reproducible apparent progress.

\end{enumerate}

\begin{figure}[t]
  \centering
  \includegraphics[width=\columnwidth]{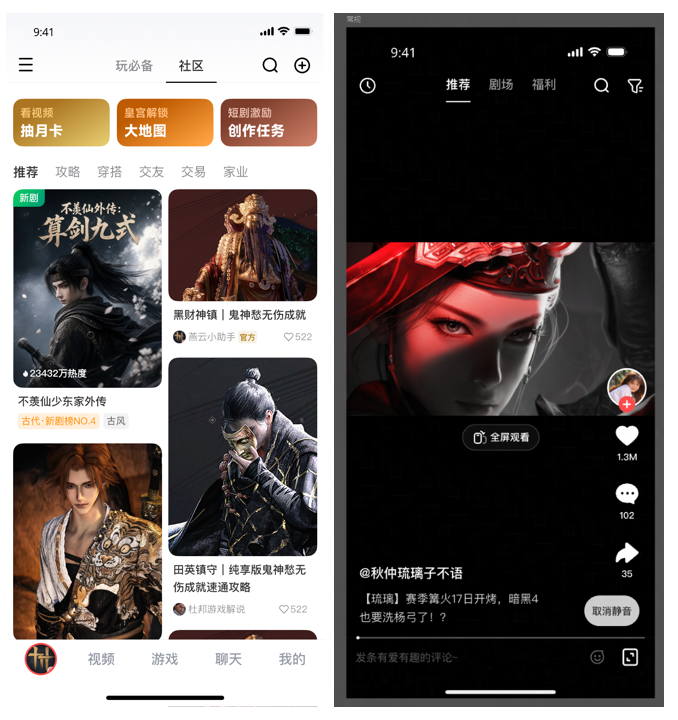}
  \caption{The two core DASHEN recommendation scenarios studied in this paper: the unified single- and two-column feed and the immersive-video feed. }
  \Description{Two DASHEN mobile screenshots side by side. The left screen shows the community feed in a two-column card layout. The right screen shows a full-screen immersive video player with engagement controls.}
  \label{fig:dashen}
\end{figure}

\begin{figure*}[t]
  \centering
  \includegraphics[width=\textwidth]{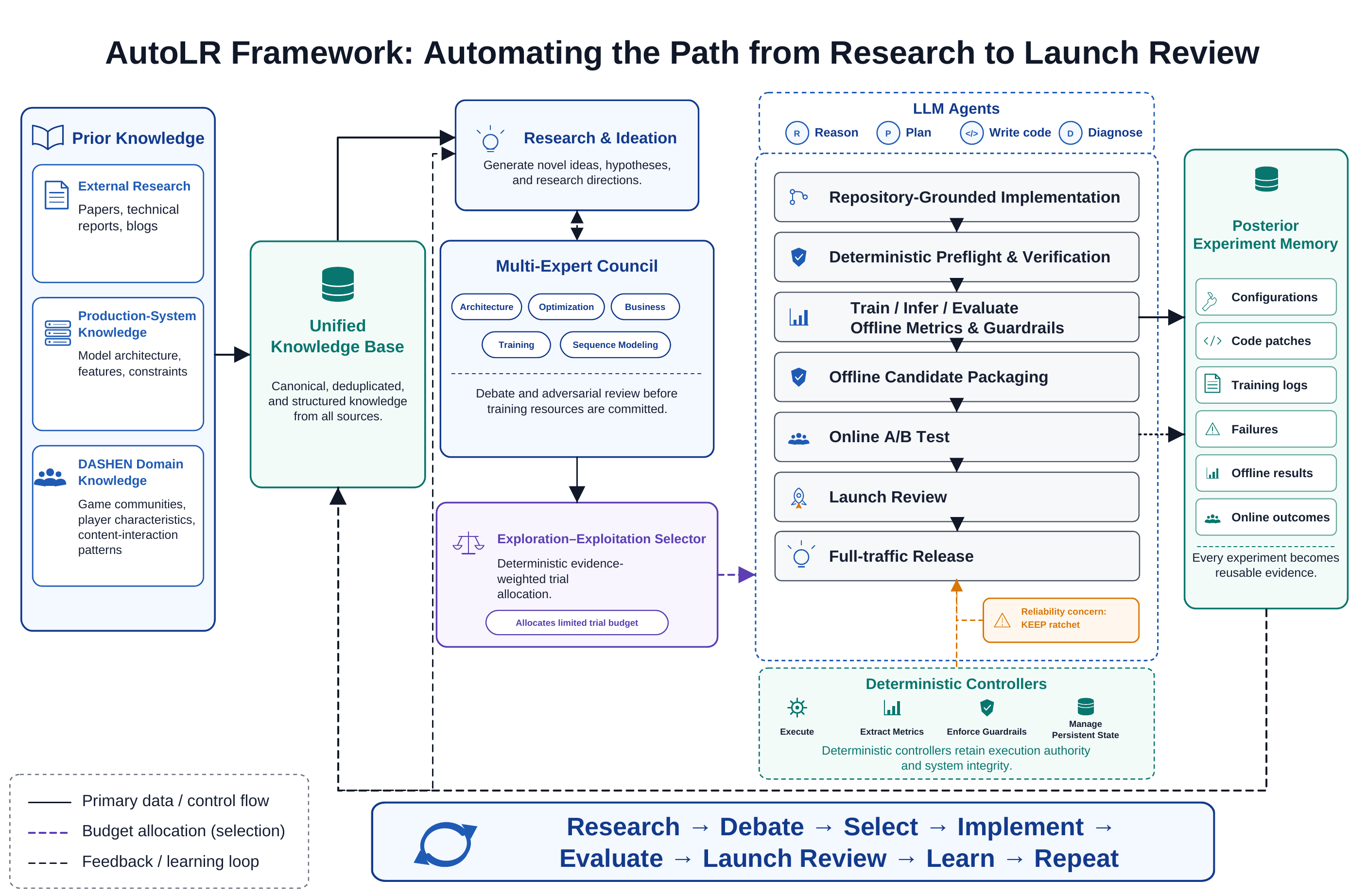}
  \caption{AutoLR architecture. Prior knowledge and experiment memory feed a multi-expert council and a budgeted selector. LLM agents propose and implement; deterministic controllers own verification, execution, metrics, guardrails, and persistent state. Autonomy ends at offline packaging; online A/B and Launch Review remain human-gated. Terminal results write back to memory---the feedback path behind the KEEP ratchet (Section~\ref{sec:keep-ratchet}).}
  \Description{A research-to-launch architecture diagram. Evidence and experiment memory feed a multi-expert council, a deterministic evidence-weighted direction selector, and repository-grounded implementation. Deterministic controllers own offline verification, execution, metrics, vector guardrails, verdicts, and candidate packaging. Online A/B testing, Launch Review, and full rollout appear in a separate dashed human-gated region, with an aspirational dashed feedback path.}
  \label{fig:overview}
\end{figure*}

Our empirical study covers two core DASHEN recommendation scenarios
(Figure~\ref{fig:dashen}): the standard recommendation feed, which supports
single- and two-column layouts, and the immersive-video feed.
Figure~\ref{fig:overview} summarizes the AutoLR lifecycle and system architecture.

\section{Related Work}
\label{sec:related}

\subsection{From AutoML to Autonomous ML Engineering}

Classical AutoML, hyperparameter optimization, and neural architecture search automate choices within a predefined search space
\citep{thornton2013autoweka,feurer2015automl,snoek2012bo,elsken2019nas,zoller2021automl}.
Recommendation-specific methods extend this paradigm to choices such as feature interactions and embedding dimensions
\citep{liu2020autofis,joglekar2020neuralinputsearch}.
In mature ranking systems, however, useful improvements are often difficult to enumerate in advance: a new sequence encoder, task relationship, sample-weighting rule, or reward formulation is not merely a parameter choice, but a hypothesis about why a particular production system should improve.

LLM agents extend automation from selecting among predefined alternatives to generating and implementing new ones. ReAct and Toolformer established widely used reasoning--acting and tool-use patterns
\citep{yao2023react,schick2023toolformer}.
MLE-bench evaluates agents on end-to-end ML-engineering tasks, while AIDE and MLE-STAR develop agents that search over and refine executable ML solutions
\citep{chan2025mlebench,jiang2025aide,nam2025mlestar}.
The AI Scientist and AlphaEvolve extend this direction toward research-hypothesis generation and evaluator-guided algorithm discovery
\citep{lu2024aiscientist,novikov2025alphaevolve}.
Multi-agent debate can broaden proposal generation and critique, although its gains over strong single-agent or ensemble baselines are mixed and depend on the debate protocol
\citep{du2023debate,smit2024mad}.

These works show that open-ended generation and iterative refinement are feasible. Industrial deployment adds a different requirement: \emph{governance}. A production system must determine what evidence is sufficient, which metrics cannot be traded away, when failures should be retried, and which components are allowed to modify persistent or production-facing state.

\subsection{Agentic Harnesses for Industrial Recommendation}

We focus on LLM agents acting as recommendation engineers that modify and evaluate ranking systems, rather than on LLMs serving directly as recommenders on the product surface.

Several recent industrial systems share this framing. Google's Self-Evolving Recommendation System combines a high-throughput offline discovery loop with slower online validation and preserves experimental history in an experiment journal
\citep{wang2026selfevolving}.
AgentX organizes proposal generation, repository-grounded implementation, guarded online A/B evaluation, and harness evolution into a closed production loop
\citep{lao2026agentx}.
RecHarness separates \emph{where to search} from \emph{how to implement}: validation-driven Thompson sampling allocates trials across predefined edit directions, while an LLM generates executable changes within the selected direction; validated candidates can then become incumbents for later rounds
\citep{ling2026recharness}.
Meta's engineering report on the Ranking Engineer Agent describes a planner--executor architecture for coordinating asynchronous ranking experiments over multi-day to multi-week workflows
\citep{kumar2026rea}.

Other systems emphasize complementary parts of the same problem. NOVA uses a verification cascade that checks semantic validity, executability, offline effectiveness, and online impact before promotion
\citep{liu2026nova}.
AgenticRecTune combines multiple optimization agents with a self-evolving skill repository
\citep{wu2026agenticrectune}.
PILOT places statistical decisions, permission checks, and persistent state commits behind deterministic services, with a controlled component responsible for lifecycle-state updates
\citep{pilotteam2026}.

AutoLR overlaps substantially with these systems. Its experiment memory is related to Google's experiment journal and AgentX's accumulated experiment knowledge; its selector is conceptually related to RecHarness's separation of direction selection from concrete mutation; and its deterministic controller follows the broader principle, also present in NOVA and PILOT, that model-generated proposals must pass machine-enforced gates. We therefore do not claim that AutoLR introduces a uniquely long automation pipeline or a unique division of authority.

Instead, AutoLR provides a Launch-Review-centered, longitudinal production study of an evidence-governed experimentation loop. In the deployment studied here, AutoLR operates autonomously from research and proposal generation through code modification, training, offline evaluation, and candidate packaging, while engineers retain authority over admission to online A/B testing and full-traffic rollout. We additionally study a reliability issue that arises when accepted offline candidates are committed as the baselines for subsequent experiments.

Long-running operation further requires explicit state, bounded context, resumable execution, failure classification, and a clear separation between model reasoning and authoritative system actions
\citep{rajasekaran2026harness,kumar2026rea,pilotteam2026}.
Section~\ref{sec:system} describes how AutoLR implements these requirements.

\subsection{Adaptive Evaluation and the Mutable Baseline}

Long-running experimentation faces three related reliability risks.

First, repeatedly choosing new hypotheses after inspecting the same held-out data can overfit the evaluation set and invalidate ordinary statistical conclusions
\citep{dwork2015reusable,russo2016bias}.
Methods such as reusable holdouts, the Ladder, always-valid inference, and online false-discovery control are designed to reduce this form of adaptive overfitting
\citep{blum2015ladder,johari2015always,javanmard2015online}.

Second, even with fresh evaluation data, selecting the best result from many noisy trials creates selection bias: the chosen candidate is more likely to have benefited from favorable noise
\citep{smith2006optimizer}.
This winner's-curse effect has also been observed in industrial online controlled experiments
\citep{lee2018winner}.
The issue becomes particularly important when reported gains are comparable to variation caused by random seeds, benchmark noise, or baseline tuning
\citep{bouthillier2021variance,reimers2017reporting,dacrema2019progress,rendle2020baselines}.

Third, an autonomous experimentation system can make such errors persistent. When a candidate is accepted and becomes the baseline for later trials, an erroneous improvement is no longer only a noisy measurement---it changes the working repository against which future candidates are generated and evaluated. We call this stateful amplification mechanism the \textbf{KEEP ratchet}. It can amplify errors caused by adaptive evaluation, noisy candidate selection, or ordinary run-to-run variance, and can arise in any iterative system that promotes accepted candidates to incumbents
\citep{ling2026recharness}.

Existing statistical methods can reduce the probability of accepting a false improvement, but they do not by themselves determine how a long-running system should commit, confirm, or roll back persistent artifacts. The Ladder provides a useful analogy because it restricts updates to a maintained reference
\citep{blum2015ladder}.
The operational consequence is stronger in AutoLR: a KEEP changes the repository artifact used for subsequent experiments rather than only updating a reported score.

Finally, recommendation optimization is inherently multi-objective
\citep{jannach2022multiobjective,kohavi2009controlled,deng2016metrics,dmitriev2017dirty}.
AutoLR therefore combines a primary optimization score with independent guardrails on protected metrics. A candidate can rank highly on the primary score and still be rejected if it violates a protected constraint. This non-compensatory rule mirrors the veto semantics of human Launch Review and forms the basis of the promotion protocol in Section~\ref{sec:system}.

\section{AutoLR System}
\label{sec:system}
\subsection{Lifecycle State and Authority}

AutoLR automates the workflow from research through offline candidate packaging. Given a business-motivated optimization request, a fixed offline evaluation contract, and repository constraints, it retrieves evidence, proposes and reviews research directions, selects experiments, modifies the repository, launches training and evaluation, diagnoses outcomes, and packages promising candidates. Admission to online A/B testing and full-traffic rollout remains human-gated.

At round $t$, let $T_t$ denote the active trunk, $K_t$ the available knowledge and experiment memory, $H_t$ the experiment history, and $B_t$ the remaining trial budget. AutoLR executes the following lifecycle:

\begin{equation}
\begin{aligned}
& a_t = S(T_t,K_t,H_t,B_t), \\
& p_t = G(a_t,T_t,K_t,H_t), \\
& \pi_t = I(p_t,T_t), \\
& (\widehat{\mathbf m}_t,\ell_t) = E(T_t\oplus\pi_t).
\end{aligned}
\label{eq:lifecycle}
\end{equation}

Here, $a_t$ is the research direction selected for round $t$; $p_t$ is the concrete, council-reviewed proposal within that direction; $\pi_t$ is the repository-grounded code patch implementing the proposal; $\widehat{\mathbf m}_t$ is the resulting offline metric vector; and $\ell_t$ contains execution evidence such as validation results, training status, checkpoints, and failure diagnostics. Correspondingly, $S$ is the direction selector, $G$ generates and reviews a concrete proposal, $I$ converts the proposal into a legal repository modification, and $E$ is the deterministic training-and-offline-evaluation protocol. The resulting metrics, execution evidence, and terminal decision are then appended to the experiment history and written back as structured experiment memory.

Authority is deliberately asymmetric. LLM agents handle literature synthesis, hypothesis generation, debate, repository understanding, code generation, and qualitative diagnosis. Deterministic controllers govern repository checks, job execution, timeouts, metric extraction, guardrails, git transitions, and persistent offline state. \textbf{LLM outputs are proposals, not state transitions}: every authoritative transition requires a machine-validated artifact such as a verified proposal, commit, checkpoint, metric record, or promotion verdict.

\subsection{Multi-Expert Proposal Council}

A recommendation optimization request typically specifies the target metrics and offline evaluation protocol, but not how the model should be improved. The design space remains broad: gains may come from model architecture, feature interaction, sample utilization, multi-objective learning, short- and long-term behavior modeling, loss design, or training strategy. AutoLR therefore converts the request into a structured task contract containing the optimization target, evaluation protocol, allowed edit surfaces, hard constraints, protected metrics, relevant evidence, and unresolved assumptions.

The \textbf{Multi-Expert Proposal Council} evaluates this task through a fixed multi-stage process. First, role-specialized agents independently analyze the same frozen briefing without seeing one another's initial proposals. Typical roles cover model architecture, feature interaction, sequence modeling, multi-task learning, loss and optimization, DASHEN business semantics, repository feasibility, and adversarial review. This independent first pass encourages diverse hypotheses and reduces early anchoring.

The proposals then enter a \textbf{roundtable synthesis} that clusters similar mechanisms, removes redundancy, identifies genuine disagreements, and strengthens competing arguments before rejection. Surviving proposals subsequently undergo \textbf{adversarial review}: dedicated challenge agents search for unsupported assumptions, implementation infeasibility, missing activation conditions, metric risks, and groupthink. High-risk proposals may trigger a rebuttal from the original specialist, supported by concrete evidence or a falsifiable validation probe.

\begin{figure*}[t]
  \centering
  \includegraphics[width=\textwidth]{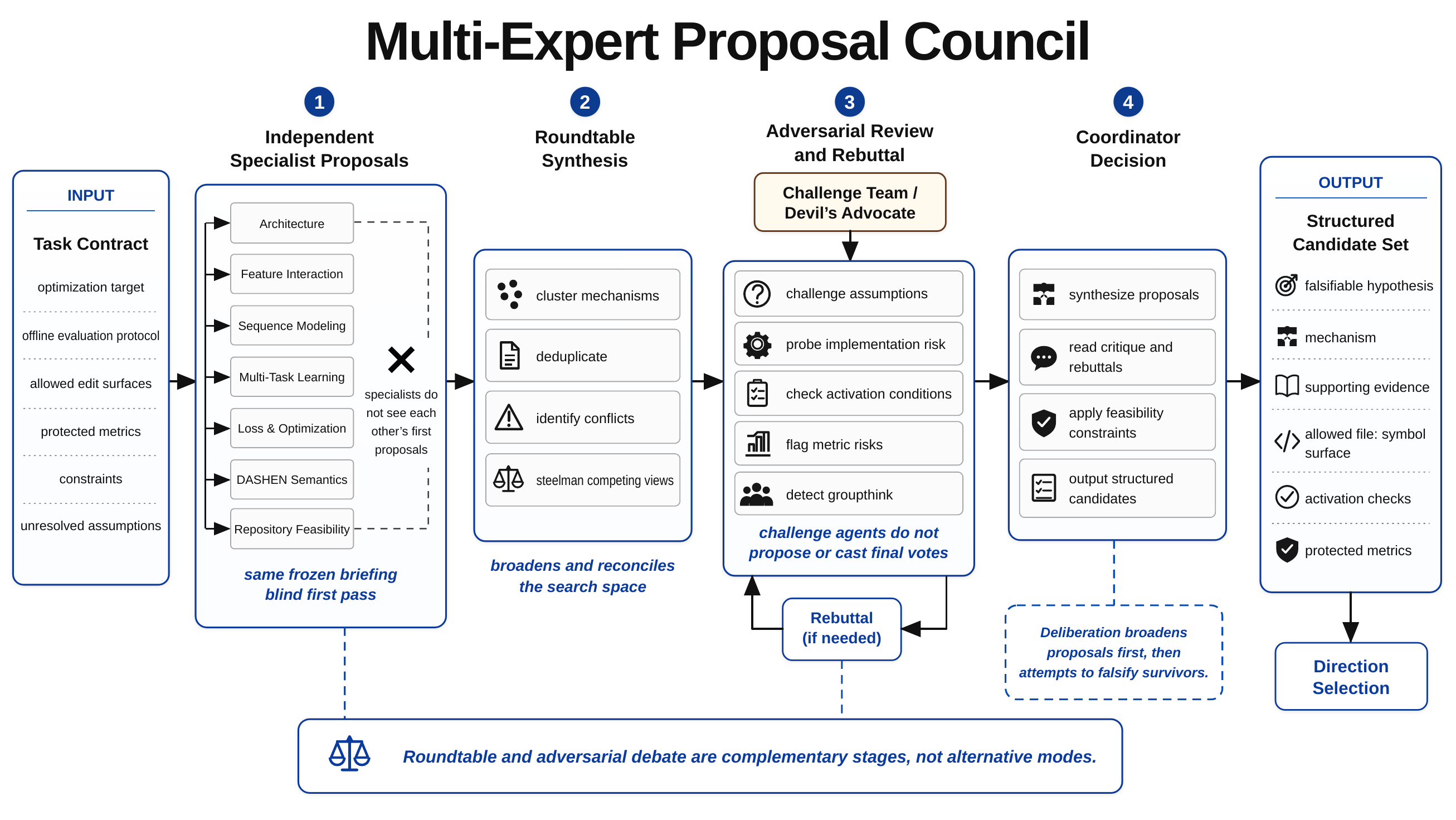}
  \caption{Four-stage Multi-Expert Proposal Council. Independent specialists propose from a frozen task contract; roundtable synthesis clusters and reconciles mechanisms; adversarial review attempts to falsify survivors, with optional rebuttal; a coordinator emits a structured candidate set with a declared edit surface and activation checks.}
  \Description{A left-to-right process diagram. A task-contract input feeds independent specialist proposals, roundtable synthesis, adversarial review with an optional rebuttal loop, and a coordinator decision, producing a structured candidate set that proceeds to direction selection.}
  \label{fig:council}
\end{figure*}

Finally, a \textbf{Coordinator} synthesizes the proposals, critiques, rebuttals, historical evidence, and feasibility constraints into a structured candidate set. Each candidate specifies a falsifiable hypothesis and mechanism, supporting evidence, the intended model or training change, the allowed \texttt{file:symbol} edit surface, expected observables and activation checks, and protected metrics. Hard constraints are non-compensatory: an infeasible proposal is rejected regardless of its novelty or expected gain. Figure~\ref{fig:council} summarizes this four-stage process.

\begin{figure*}[t]
  \centering
  \includegraphics[width=\textwidth]{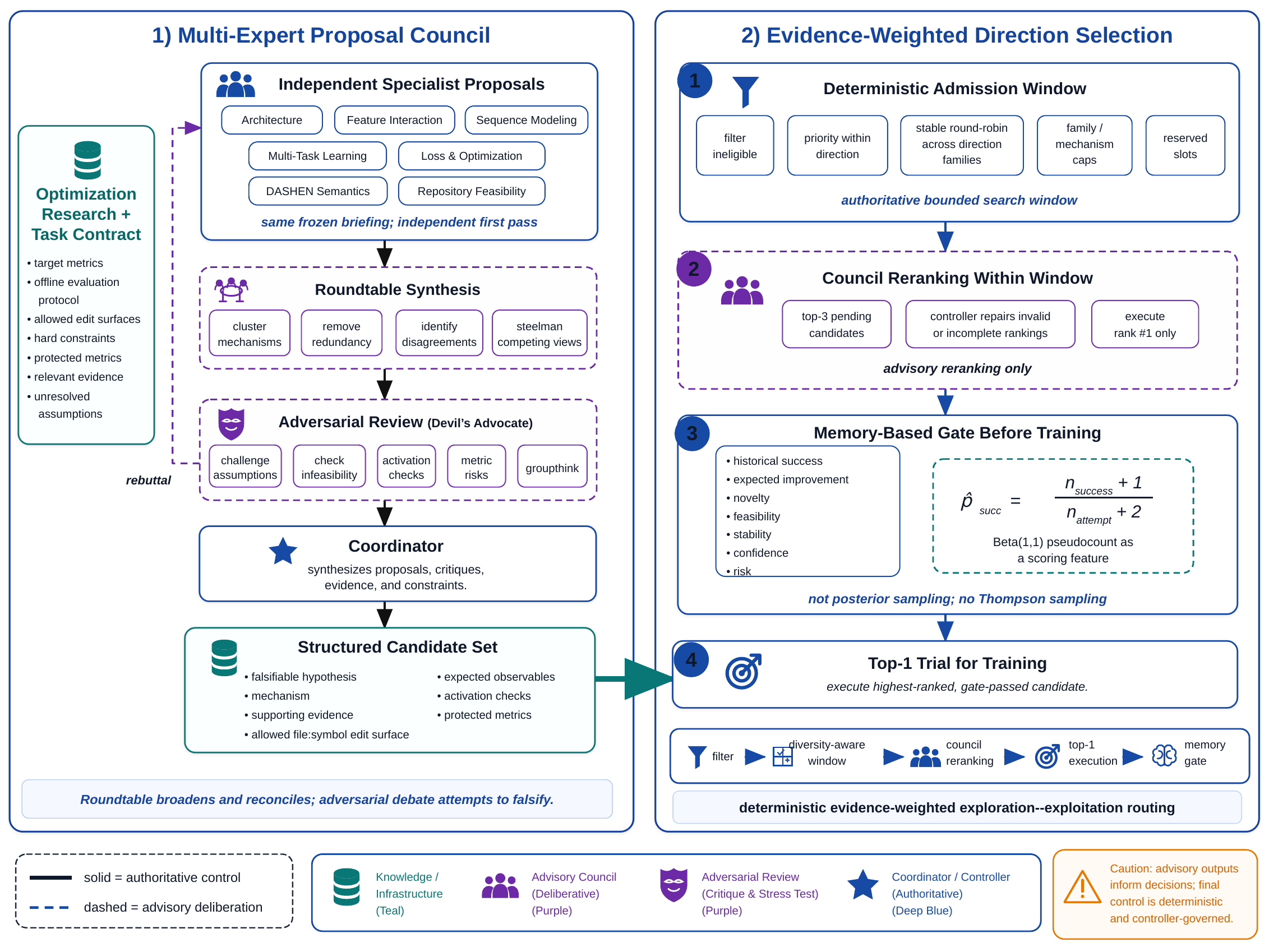}
  \caption{Pipeline connecting the Multi-Expert Proposal Council to evidence-weighted direction selection. Council deliberation is advisory; admission, top-1 execution, and the memory-based gate remain controller-governed.}
  \Description{A two-panel pipeline. The left panel shows council stages from a task contract to a structured candidate set. The right panel shows deterministic admission, council reranking, a memory-based gate, and top-1 training. Solid arrows mark authoritative control and dashed arrows mark advisory deliberation.}
  \label{fig:council-selector}
\end{figure*}

Thus, the roundtable and adversarial debate are complementary stages rather than alternative modes: the former broadens and reconciles the search space, while the latter attempts to falsify the surviving proposals. The resulting candidates are passed to the direction-selection stage, which determines where the next experiment budget should be allocated (Figure~\ref{fig:council-selector}).

\subsection{Evidence-Weighted Direction Selection}

Let $\mathcal{A}={a_1,\ldots,a_K}$ denote interpretable optimization directions such as sequence modeling, expert routing, feature interaction, gating, task weighting, regularization, and training strategy. Direction selection determines \emph{where to search}; once a direction is admitted, the council and implementation agents determine \emph{what concrete change to propose and implement} within that search region.

\begin{figure*}[t]
  \centering
  \includegraphics[width=\textwidth]{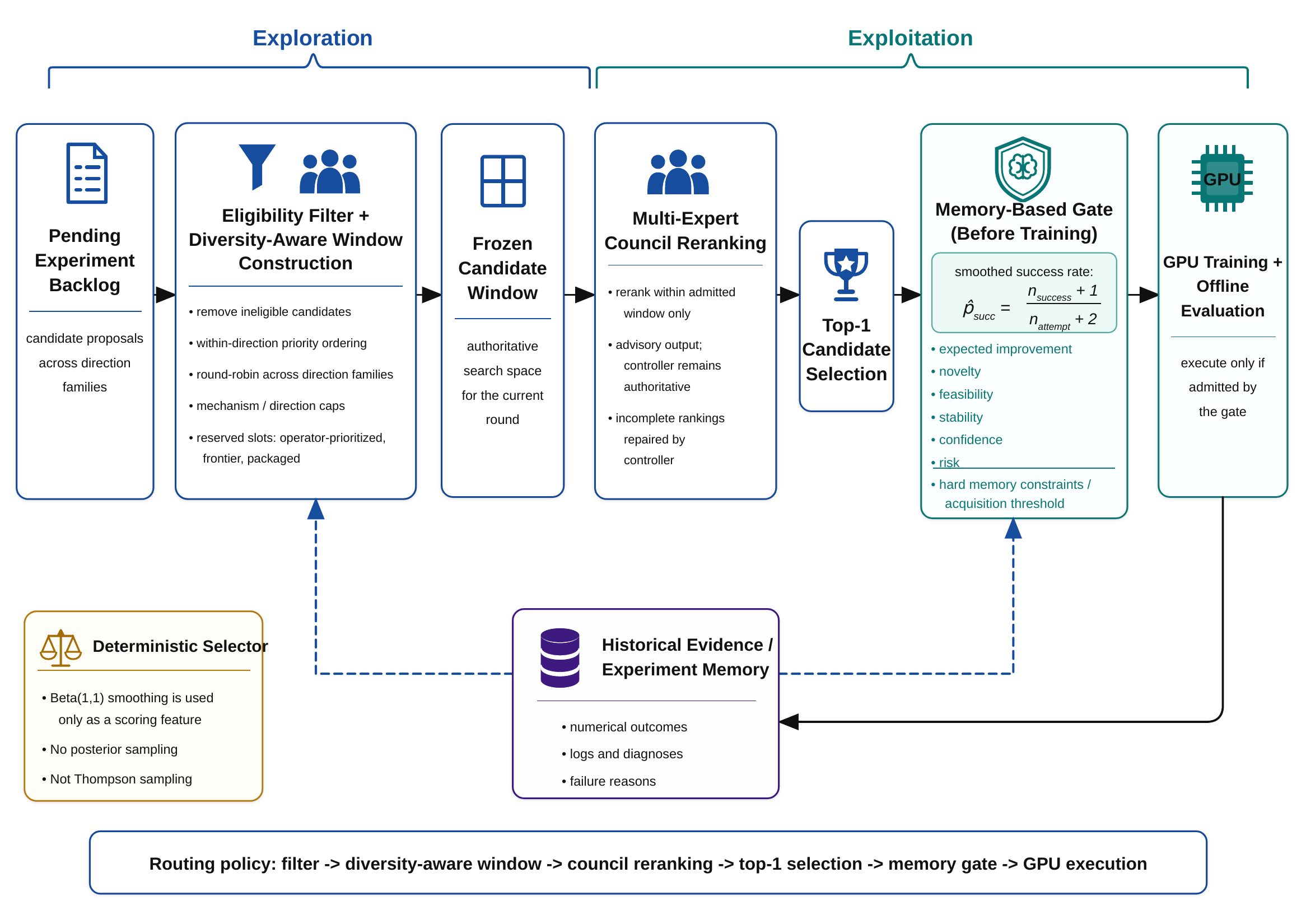}
  \caption{Evidence-weighted direction selection. A diversity-aware frozen window defines the search space; the council reranks within that window; a memory-based gate scores candidates using historical evidence before GPU training. Numerical outcomes write back to experiment memory.}
  \Description{A left-to-right routing diagram. A pending experiment backlog is filtered into a frozen candidate window, then council-reranked, gated by historical memory, and executed as a top-1 GPU training trial. Dashed arrows return outcomes to experiment memory, which feeds the filter and the memory-based gate.}
  \label{fig:selector}
\end{figure*}

The production selector is deterministic and operates in three stages (Figure~\ref{fig:selector}). First, the controller constructs a bounded candidate window from the pending experiment backlog. Ineligible candidates are removed, proposals within each direction are ordered by priority, and a stable round-robin schedule distributes slots across direction families. Additional mechanism- and direction-level caps prevent the window from being dominated by a single family, while reserved slots preserve operator-prioritized, frontier, and packaged candidates. This frozen window defines the authoritative search space for the current round.

Second, the Multi-Expert Council reranks candidates \emph{within} this window. It returns the most promising pending candidates in priority order, after which the controller moves them to the front of the queue and executes only the top-ranked candidate. Council output is advisory rather than authoritative: candidates outside the admitted window cannot be selected, incomplete rankings are repaired by the controller, and operator-prioritized candidates retain precedence.

Third, a memory-based gate is applied before training. Candidate quality is scored using deterministic features derived from prior experiments, including smoothed historical success rate, expected improvement, novelty, feasibility, stability, confidence, and risk. The historical success term uses a Beta$(1,1)$ pseudocount,

$$
\widehat{p}_{\mathrm{succ}}
=
\frac{n_{\mathrm{success}}+1}
     {n_{\mathrm{attempt}}+2},
$$

to avoid over-weighting directions with very few observations. This quantity is only a scoring feature; AutoLR does not sample from a Beta posterior or maintain direction-level posterior state. Candidates blocked by hard memory constraints or falling below the acquisition threshold are rejected or reselected before GPU training begins.

The resulting routing policy can therefore be summarized as:
\begin{equation*}
\begin{aligned}
&\text{filter}
\rightarrow
\text{diversity-aware window construction} \\
&\rightarrow
\text{council reranking}
\rightarrow
\text{top-1 execution}
\rightarrow
\text{memory gate}.
\end{aligned}
\end{equation*}

Numerical outcomes update the historical evidence used in later rounds, while logs, diagnoses, and failure reasons condition subsequent proposal generation.

Accordingly, we characterize AutoLR as a \textbf{deterministic, evidence-weighted direction selector with diversity constraints and council reranking}, rather than as Thompson sampling or a Bayesian bandit. The production corpus evaluates this selector only as part of the integrated AutoLR loop; it does not establish superiority over random, greedy, or direct LLM-based routing under a matched trial budget.

\subsection{Layered Knowledge and Experiment Memory}

AutoLR grounds proposal generation in three relatively stable knowledge sources and one evolving experiment-memory store. At experiment round $t$, we denote the knowledge available to the system as: 

\begin{equation}
K_t =
K^{\mathrm{ext}}
\cup
K^{\mathrm{sys}}
\cup
K^{\mathrm{dashen}}
\cup
M_t^{\mathrm{exp}}
\label{eq:knowledge}
\end{equation}

where $K_t$ denotes the knowledge available to AutoLR at round $t$; $K^{\mathrm{ext}}$ contains external research knowledge; $K^{\mathrm{sys}}$ contains facts and constraints about the production recommendation system; $K^{\mathrm{dashen}}$ contains DASHEN-specific domain knowledge; and $M_t^{\mathrm{exp}}$ is the experiment memory accumulated up to round $t$. The union in Equation~\ref{eq:knowledge} is conceptual: these sources remain separately indexed and are retrieved selectively rather than concatenated into a single context.

\begin{itemize}
\item \textbf{External research knowledge ($K^{\mathrm{ext}}$):}
papers, industrial reports, technical blogs, and reusable mechanisms related to model architecture, feature interaction, sequence modeling, multi-task learning, loss design, and optimization.

\item \textbf{Production-system knowledge ($K^{\mathrm{sys}}$):}
the current model architecture, feature contracts, repository structure, code boundaries, training and evaluation protocols, and serving constraints. This layer prevents proposals that are theoretically plausible but incompatible with the production system.

\item \textbf{DASHEN domain knowledge ($K^{\mathrm{dashen}}$):}
game-community structure, player behavior patterns, content semantics, interaction patterns, business objectives, and protected metrics. This layer helps connect generic modeling ideas to the characteristics and objectives of the target recommendation scenarios.

\item \textbf{Experiment memory ($M_t^{\mathrm{exp}}$):}
prior hypotheses, proposals, code patches, execution logs, failures, offline results, promotion decisions, and their associated lineage. Unlike the other three layers, this store evolves after every experiment and records not only what worked, but also what was attempted, what failed, and why.

\end{itemize}

\begin{figure*}[t]
  \centering
  \includegraphics[width=\textwidth]{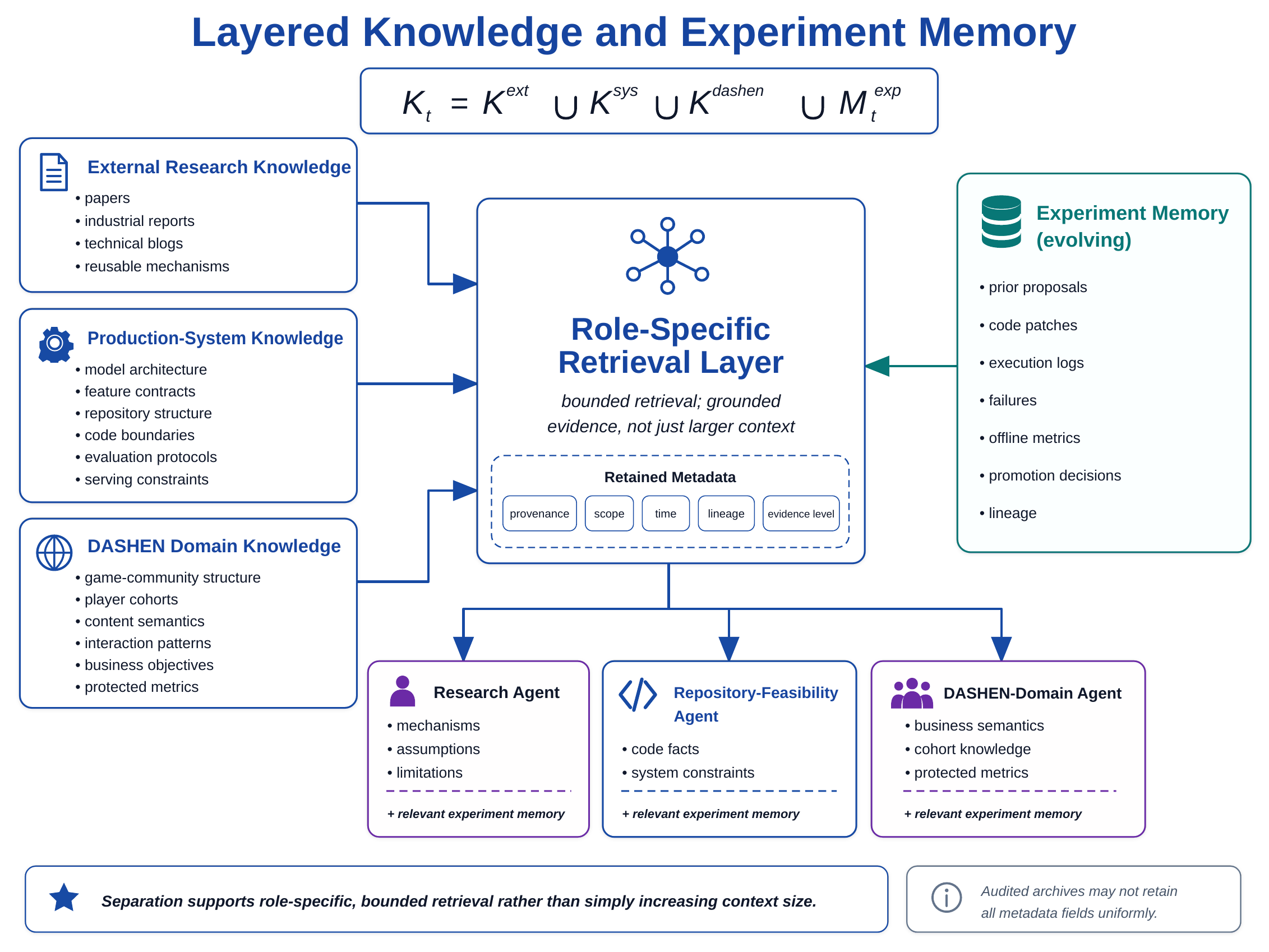}
  \caption{Layered knowledge and experiment memory. External research, production-system facts, and DASHEN domain knowledge are relatively stable; experiment memory evolves after each trial. Role-specific retrieval supplies bounded evidence to research, repository-feasibility, and domain agents rather than concatenating all sources into one context.}
  \Description{A layered architecture diagram. Four knowledge stores---external research, production-system knowledge, DASHEN domain knowledge, and evolving experiment memory---feed a role-specific retrieval layer that supplies bounded evidence to a research agent, a repository-feasibility agent, and a DASHEN-domain agent.}
  \label{fig:knowledge}
\end{figure*}

The purpose of this separation is not to maximize context size. Retrieval is role-specific and bounded (Figure~\ref{fig:knowledge}). Research agents primarily receive relevant mechanisms and limitations from prior work; repository-feasibility agents receive code and system facts; DASHEN-domain agents receive business and behavioral knowledge; and the council retrieves experiment memory to identify previously explored directions, known failure modes, and unresolved hypotheses. This reduces context dilution and helps prevent repeated exploration of equivalent ideas.

Retained items may additionally carry provenance, scope, time, lineage, and evidence level, allowing agents to distinguish, for example, a mechanism suggested by external research from an observation obtained on the current production model. Section~\ref{sec:evaluation} describes which of these fields are recoverable from the audited archive.

In short, the three knowledge layers describe \emph{what is known about the research space, production system, and application domain}, while experiment memory records \emph{what AutoLR has already tried and learned}. Together they provide the evidence used by the council and direction selector in subsequent experiment rounds.

\subsection{Constraint-Grounded Implementation and Execution}

  AutoLR treats implementation as constrained code modification rather than unconstrained generation. Before editing begins, each
  proposal specifies its intended mechanism, allowed edit surface, expected observables, and activation checks.

  A read-only preflight stage verifies that referenced files and symbols exist, required features are available, and frozen
  objectives or evaluation logic are not modified. Proposals that fail these checks are returned for revision before any long-
  running training job is launched.

  The implementation agent then modifies only the declared code surface and produces machine-checkable evidence that the intended
  mechanism has been inserted and activated. Deterministic infrastructure subsequently runs smoke tests, training, inference, metric
  extraction, and offline evaluation under explicit timeouts.

  Failures are handled according to their cause. Semantic failures, such as an unsupported assumption or inactive mechanism, return
  to proposal generation or implementation. Reproducible code or configuration failures are not retried unchanged, while transient
  infrastructure failures receive bounded retries. Every terminal outcome, including failures, is recorded as a structured
  experiment event so that later agents can distinguish an ineffective hypothesis from an execution failure.

  \subsection{Offline Promotion and Launch-Review Handoff}

  AutoLR does not directly optimize the online business metrics used in Launch Review. Each experiment session uses a fixed offline evaluation contract defined by engineers. In the deployments studied here, click- and watch-time-related offline metrics are combined into a scalar proxy, while online A/B tests evaluate downstream business outcomes such as content-consumption penetration and total consumption time.
  
  Let $\mathbf{m}(T)$ denote the offline metric vector for repository snapshot $T$. The scalar score and candidate improvement are
  
  \begin{equation}
  \begin{aligned}
  s(T)
  &=
  \mathbf{w}^{\top}\phi\left(\mathbf{m}(T)\right), \\
  \Delta_{s,t}
  &=
  s(T_t \oplus \pi_t)-s(T_t),
  \end{aligned}
  \label{eq:score}
  \end{equation}
  
  where $\phi(\cdot)$ aligns and normalizes the constituent metrics, $\mathbf{w}$ is fixed by the session-level evaluation contract, and $\Delta_{s,t}$ measures the improvement of candidate patch $\pi_t$ over the active trunk $T_t$.
  
  Promotion also requires all protected metrics to satisfy their guardrails. Let $\mathcal{G}$ be the protected metric set, $q_k \in \{-1,+1\}$ indicate the preferred direction of metric $k$, and $\epsilon_k \geq 0$ its allowed regression. Candidate feasibility is
  
  \begin{equation}
  \mathcal{F}_t
  =
  \mathbf{1}\left[
  q_k
  \left(
  m_k(T_t \oplus \pi_t)-m_k(T_t)
  \right)
  \geq -\epsilon_k,
  \quad
  \forall k \in \mathcal{G}
  \right].
  \label{eq:feasibility}
  \end{equation}
  
  Given an evidence-complete evaluation, the controller assigns a terminal action
  \begin{equation}
  d_t=
  \begin{cases}
  \textsc{Discard}, & \mathcal{F}_t=0,\\
  \textsc{Keep}_{E1}, & \mathcal{F}_t=1,\ \Delta_{s,t}\ge\tau_{\mathrm{keep}},\\
  \textsc{Pack}, & \mathcal{F}_t=1,\ \tau_{\mathrm{pack}}\le\Delta_{s,t}<\tau_{\mathrm{keep}},\\
  \textsc{Discard}, & \text{otherwise}.
  \end{cases}
  \label{eq:verdict}
  \end{equation}
  The $E1$ suffix marks a one-run offline threshold crossing rather than a confirmed effect. KEEP is the only action that rewrites the working repository; PACK retains the candidate and its lineage, and DISCARD rejects it:
  \begin{equation}
  T_{t+1}=
  \begin{cases}
  T_t\oplus\pi_t, & d_t=\textsc{Keep}_{E1},\\
  T_t, & \text{otherwise}.
  \end{cases}
  \label{eq:trunk-update}
  \end{equation}
  Equation~\eqref{eq:verdict} is a decision contract for rows with complete scalar and guardrail evidence. Missing historical payload is not treated as compliance.
  
  The retained controller also implements an optional variance-aware KEEP edge
  \begin{equation}
  \tau_t^{\mathrm{impl}}
  =
  \begin{cases}
  \max\!\left(
  \tau_{\mathrm{keep}},\,
  \lambda_{\sigma}\widehat{\sigma}_t
  \right),
  & \text{if a profile is available}, \\[2pt]
  \tau_{\mathrm{keep}},
  & \text{otherwise}.
  \end{cases}
  \label{eq:implemented-threshold}
  \end{equation}
  where $\widehat{\sigma}_t$ summarizes run-to-run variation from a compatible repeat-evaluation profile. When that profile is present, $\tau_{\mathrm{keep}}$ in Equation~\eqref{eq:verdict} is replaced by $\tau_t^{\mathrm{impl}}$.
  
  In the audited configuration, noise calibration was disabled and no reusable version-matched profile was retained; the controller therefore used the fixed floor $\tau_{\mathrm{keep}}$ rather than blocking KEEP. Section~\ref{sec:results} analyzes this runtime gap. A safer future protocol would defer persistent promotion when no compatible calibration profile is available and obtain protocol-matched repeat evaluations before committing a new trunk. Such a condition could be represented by a non-promotable \texttt{CALIBRATION\_REQUIRED} state; this is a proposed safeguard, not an implemented state in the audited controller.
  
  Candidates that pass offline review are packaged with their code, metrics, lineage, and supporting evidence. Engineers currently decide which packages enter online A/B testing, and Launch Review remains the human gate for full-traffic rollout. Because the audited archive does not retain complete immutable linkage between offline candidates and online packages, offline and online evidence are analyzed separately.
  
  \section{Production Evaluation}
  \label{sec:evaluation}
  
  \subsection{Setting and Corpus}
  
  The audited corpus covers two DASHEN recommendation scenarios: the single- and two-column feed and the immersive-video feed. Both use fixed multi-objective offline evaluation protocols. Within an experiment session, AutoLR operates within a constrained edit scope: it may modify designated model and configuration code, while the training and evaluation pipeline remains fixed. Architecture-level changes are introduced as new, separately validated modules rather than by overwriting existing production code.
  
\begin{figure*}[t]
      \centering
      \includegraphics[width=\textwidth]{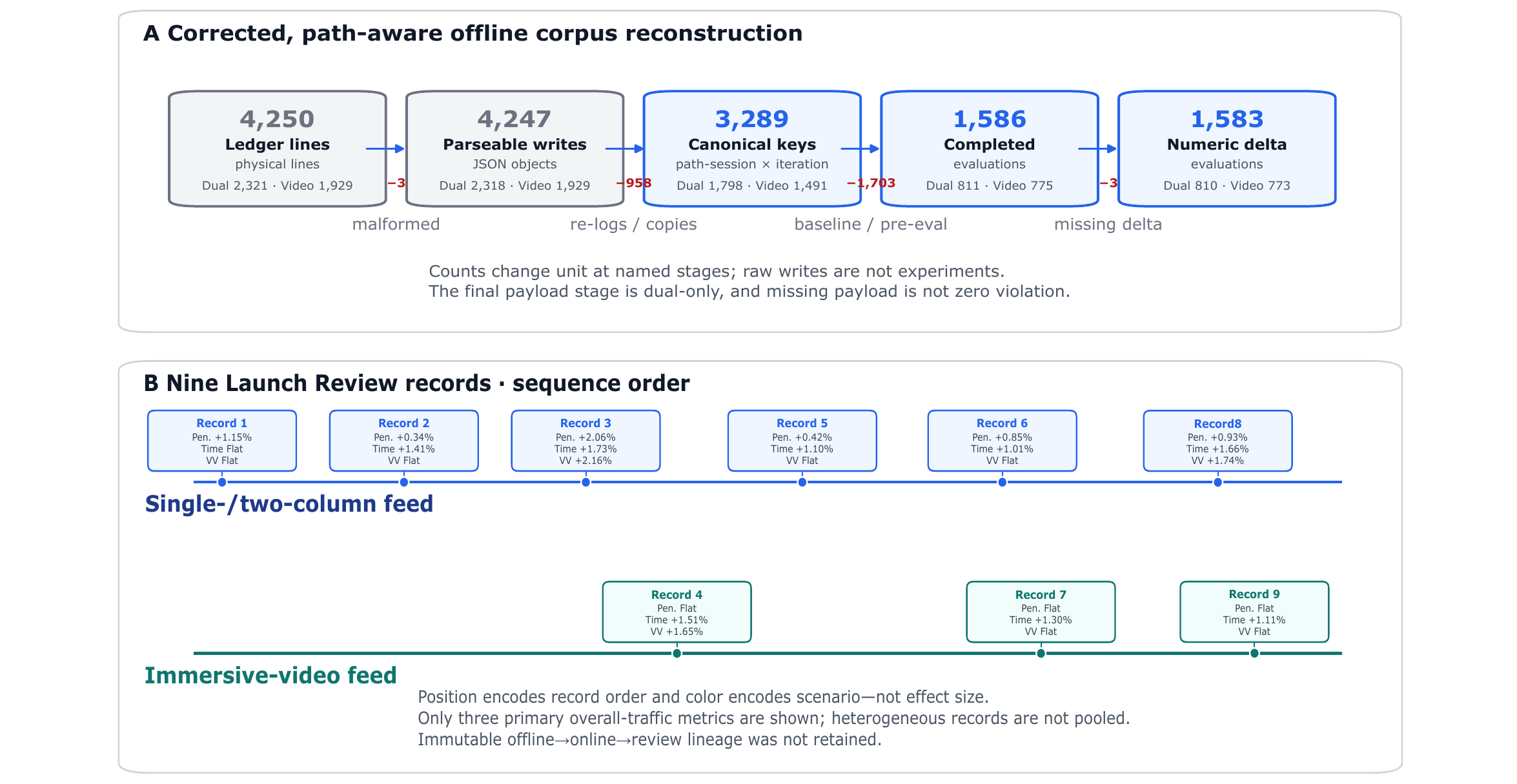}
      \caption{Production evidence. (A) Path-aware reconstruction reduces 4,250 physical ledger records to 1,586 completed evaluations, of which 1,583 retain numeric scalar deltas. (B) Nine separately maintained production Launch Review records are shown in sequence order for the two recommendation scenarios and restricted to three primary overall-traffic metrics. The records are heterogeneous and are not pooled; immutable offline-to-online-to-review linkage is not retained in the audited repositories.}
      \Description{A two-panel figure. Panel A shows the corpus reconstruction from 4,250 physical ledger records through 4,247 parseable records and 3,289 canonical experiment records to 1,586 completed evaluations and 1,583 numeric scalar deltas. Panel B shows nine Launch Review records---six for the single- and two-column feed and three for the immersive-video feed---reporting content-consumption penetration rate, total content-consumption time, and valid content views when available.}
      \label{fig:production-evidence}
  \end{figure*}

  The audit reconstructs the offline experiment corpus from 4,250 non-empty ledger records, of which 4,247 are parseable. Path-aware deduplication keyed by scenario, path session, and iteration produces 3,289 canonical experiment records. A session is identified from the enclosing archive path and groups a sequence of related experiment iterations. After filtering to completed non-baseline evaluations, the final corpus contains 1,586 completed evaluations; 1,583 retain a numeric scalar delta versus the active trunk and are used in the promotion-resolution analysis. Figure~\ref{fig:production-evidence} summarizes the reconstruction flow together with the available production evidence.
  
  Evaluations within the same experiment session are correlated because they may share data windows, active trunks, and experiment history. We therefore compute uncertainty intervals by resampling whole sessions within each recommendation scenario, rather than treating individual evaluation rows as independent samples.
  \subsection{Evaluation Questions}
  
  We organize the evaluation around three descriptive questions:
  
  \begin{itemize}
      \item \textbf{RQ1:} At what scale did AutoLR support routine model iteration under controller-governed execution, and how were human responsibilities and failures handled?

      \item \textbf{RQ2:} What online movements were reported for engineer-selected packages in the available production Launch Review records?
  
      \item \textbf{RQ3:} What reliability risks arise when a one-run offline result can change the baseline used by subsequent experiments?
  \end{itemize}
  
  The LLM-cost analysis is reported separately as an operational observation rather than a research question. Harness version, model family, task mix, and accumulated experiment memory changed together during the migration, so the historical corpus does not support causal attribution to any single factor.
  
\section{Results}
\label{sec:results}

\subsection{RQ1: Operating Scale and Routine Production Use}

AutoLR completed 1,586 evaluations across the single- and two-column feed and the immersive-video feed (Figure~\ref{fig:production-evidence}A). In the two studied scenarios, AutoLR has become the primary workflow for routine model iteration: it carries out proposal execution, repository-grounded implementation, training, offline evaluation, and experiment-state updates, while deterministic services control the train--infer--evaluate critical path.

AutoLR changes the role of algorithm engineers rather than removing it. Engineers define optimization objectives, evaluation contracts, protected metrics, and code boundaries; maintain production and domain knowledge; and supervise the autonomous loop. Routine research, implementation, training, and offline evaluation are delegated to AutoLR, while engineers intervene on ambiguous failures or changing business requirements and retain authority over online A/B admission and full-traffic rollout. In this model, human effort shifts from repeated execution toward direction setting, evidence quality, system governance, and production decisions.

The archive also records structured failures and recovery events, including infeasible proposals, inactive mechanisms, smoke-test failures, timeouts, and training crashes. Because one candidate may generate multiple retry or recovery events, these records are used to characterize failure modes rather than estimate a candidate-level failure rate. Together with the 1,586 completed evaluations, they show that AutoLR supports sustained production experimentation with explicit failure handling. This evidence demonstrates operational robustness, but not the superiority of individual components such as the council or direction selector.

\subsection{RQ2: Production Launch Review Outcomes}

AutoLR has become a stable part of routine model iteration in the single- and two-column feed and the immersive-video feed. Nine production Launch Review records are available from these two scenarios, providing evidence that AutoLR-generated candidates progressed beyond offline evaluation into online A/B testing and production launch decisions. 

Table~\ref{tab:online} summarizes the reported online movements. Positive relative changes were observed in content-consumption penetration, total content-consumption time, and valid content views across the nine records; metrics not separately disclosed in a review are treated as flat. Under this convention, the arithmetic sums of the relative-lift values are +5.75\%, +10.83\%, and +5.55\%, respectively. 

Beyond the two primary scenarios, AutoLR has also been transferred to DASHEN search, user-growth, and commercialization workflows. Search and user-growth have each completed a Launch Review with positive reported online results. In commercialization, AutoLR has completed initial exploration and produced clear offline AUC (Area Under the Curve) improvements, with online validation currently in progress. These additional deployments are part of an ongoing expansion of AutoLR across DASHEN. We report the current progress briefly here; a more comprehensive evaluation will follow as additional scenarios complete online validation.

Taken together, the evidence shows that AutoLR is not limited to an offline experimentation prototype: it has been incorporated into routine production iteration in the two core recommendation scenarios and has begun to generalize to additional ranking tasks. The cross-scenario results support the practical deployability and broader production potential of the harness, while stronger claims about transfer effectiveness will require complete lineage and scenario-specific online evaluation.

\begin{table*}[t]
\caption{Nine production Launch Review records in DASHEN's two primary recommendation scenarios, indexed by sequence. Values are reported relative changes against contemporaneous controls. Only the three primary overall-traffic metrics are shown; Flat denotes no reported overall-traffic movement.}
\label{tab:online}
\centering
\footnotesize
\setlength{\tabcolsep}{4pt}
\begin{tabularx}{\textwidth}{@{}c X c c c@{}}
\toprule
No.
& Scenario
& \shortstack{Content-consumption\\penetration rate}
& \shortstack{Total content-\\consumption time}
& \shortstack{Valid content\\views (VV)} \\
\midrule
1 & Single-/two-column feed & +1.15\% & Flat & Flat \\
2 & Single-/two-column feed & +0.34\% & +1.41\% & Flat \\
3 & Single-/two-column feed & +2.06\% & +1.73\% & +2.16\% \\
4 & Immersive-video feed & Flat & +1.51\% & +1.65\% \\
5 & Single-/two-column feed & +0.42\% & +1.10\% & Flat \\
6 & Single-/two-column feed & +0.85\% & +1.01\% & Flat \\
7 & Immersive-video feed & Flat & +1.30\% & Flat \\
8 & Single-/two-column feed & +0.93\% & +1.66\% & +1.74\% \\
9 & Immersive-video feed & Flat & +1.11\% & Flat \\
\bottomrule
\end{tabularx}
\end{table*}

\begin{figure*}[t]
    \centering
    \includegraphics[width=\textwidth]{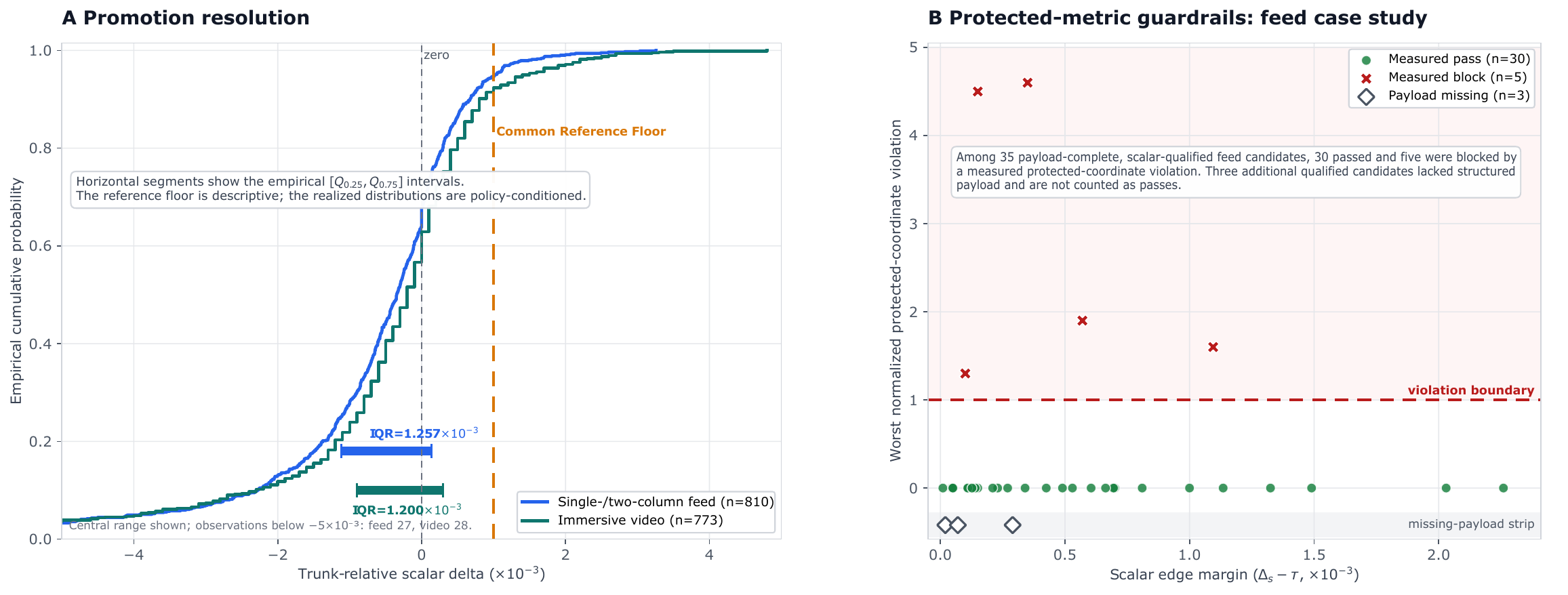}
    \caption{Promotion reliability. (A) The historical distribution of trunk-relative scalar deltas is compared with the common reference floor $10^{-3}$; this is a descriptive replay rather than proof of an operative threshold for rows without threshold provenance. (B) The available single-/two-column guardrail records illustrate that scalar-qualified candidates can still be blocked by protected-metric violations; missing structured payload is shown separately and is not interpreted as compliance.}
    \Description{A two-panel figure. Panel A shows empirical distributions of trunk-relative scalar deltas for the single- and two-column feed and the immersive-video feed together with the common reference floor. Panel B shows the available single-/two-column guardrail case study, separating measured protected-metric violations from missing structured payloads.}
    \label{fig:reliability}
\end{figure*}

\subsection{RQ3: Reliability of Offline Promotion}
  
\paragraph{Promotion resolution and threshold provenance.}

Because historical threshold provenance is incomplete, we replay a common
reference floor of $\tau_{\mathrm{ref}}=10^{-3}$ rather than treating it as
the audited operative threshold for every evaluation. For each scenario, we
compute the empirical interquartile range as
\[
\operatorname{IQR}(\Delta_s)
=
Q_{0.75}(\Delta_s)-Q_{0.25}(\Delta_s),
\]
over all completed non-baseline evaluations with a numeric trunk-relative
scalar delta. This gives $1.257\times10^{-3}$ over 810 deltas in the
single-/two-column feed and $1.201\times10^{-3}$ over 773 deltas in the
immersive-video feed (Figure~\ref{fig:reliability}A). The reference floor is therefore comparable in
magnitude to the central spread of the realized candidate deltas. It is
crossed by 4.7\% and 8.5\% of numeric evaluations, while the observed KEEP
fractions are 3.82\% and 5.03\%, respectively
(Table~\ref{tab:resolution}).

\begin{table}[t]
\caption{Promotion-resolution statistics. Crossing rates use the common
reference floor $\tau_{\mathrm{ref}}=10^{-3}$ for descriptive replay; they
are not interpreted as audited operative-threshold rates for evaluations
without threshold provenance.}
\label{tab:resolution}
\centering
\small
\setlength{\tabcolsep}{4pt}
\begin{tabular}{@{}lrr@{}}
\toprule
Quantity
& \shortstack{Single-/\\two-column}
& \shortstack{Immersive\\video} \\
\midrule
Numeric scalar deltas
& 810
& 773 \\
$\operatorname{IQR}(\Delta_s)$
& $1.257\times10^{-3}$
& $1.201\times10^{-3}$ \\
Reference-floor crossing
& 4.7\%
& 8.5\% \\
\midrule
Observed KEEP fraction
& 3.82\%
& 5.03\% \\
Observed PACK fraction
& 12.58\%
& 7.35\% \\
Recorded threshold provenance
& 480
& 0 \\
\bottomrule
\end{tabular}
\end{table}

Reference-floor crossing rates use only evaluations with numeric scalar
deltas, whereas KEEP and PACK fractions use all completed evaluations in
the corresponding scenario. These statistics indicate that the reference
floor is selective relative to the realized candidate distribution.
However, the IQR summarizes the dispersion of heterogeneous candidate
outcomes rather than protocol-specific run-to-run noise. It therefore does
not establish statistical significance, an optimal threshold, or the
operative historical threshold for evaluations without provenance.
Moreover, the observed delta distribution is policy-conditioned because
each accepted KEEP changes the trunk against which subsequent candidates
are evaluated.

\paragraph{Calibration gap and mutable-baseline risk.}
The retained controller contains an optional variance-aware KEEP edge, but noise calibration was disabled in the audited configuration and no reusable version-matched calibration profile was retained. The controller therefore fell back to the fixed floor rather than blocking promotion. In this configuration, a one-run threshold crossing could update the experimental trunk without a protocol-matched confirmation run.

Because KEEP rewrites the active trunk (Equation~\eqref{eq:trunk-update}), a non-reproducible acceptance does more than add one noisy result: it changes the reference used by subsequent experiments. The archive establishes this mutable-baseline mechanism, but it cannot identify how much observed progress or drift reflects true model improvement, changing candidate quality, diminishing returns, or accumulated selection noise. We refer to this stateful amplification risk as the \textbf{KEEP ratchet} and discuss its design implications in Section~\ref{sec:keep-ratchet}.

\paragraph{Protected-metric guardrails.}
Structured guardrail payloads are recoverable only for a subset of the single- and two-column feed archive, and equivalent structured payloads are not retained for the immersive-video feed. Within the recoverable subset, Figure~\ref{fig:reliability}B shows that scalar-qualified candidates can still be blocked by protected-metric violations. Missing payload is reported separately and is not interpreted as compliance. The guardrail evidence is therefore a scenario-specific case study rather than a corpus-wide estimate.

\section{Discussion and Limitations}
\label{sec:discussion}

\subsection{Implications for Governed Autonomous Experimentation}
\label{sec:keep-ratchet}

AutoLR suggests that the durable unit of autonomy is the governed experiment
lifecycle rather than any particular agent or foundation model. Deterministic
evidence contracts, guardrails, recovery policies, and state transitions allow
semantic components to evolve without granting them authority over persistent
state. The production corpus supports the operational viability of this
integrated design, but does not isolate the causal contribution of its
individual mechanisms.

The KEEP ratchet exposes a fundamental asymmetry between exploration and
commitment. Failed trials are bounded and recoverable, whereas KEEP changes the
baseline for subsequent experiments. The common reference floor is selective relative to the realized candidate
distribution, but is neither statistically calibrated nor shown to be optimal. Persistent promotion should therefore
require a practically meaningful improvement, protected-metric compliance, and
protocol-matched confirmation; when compatible calibration evidence is
unavailable, promotion should fail closed. PACK can retain promising candidates
without changing the trunk. Because online A/B admission and full-traffic
rollout remain human-gated, the immediate risk is degraded search validity and
wasted trial budget rather than automatic production deployment. These
safeguards remain design recommendations rather than evaluated components.

\subsection{Limitations}

This study demonstrates sustained production experimentation and identifies a
mutable-baseline risk, but does not estimate component-level causal effects or
a false-promotion rate. The archive lacks matched-budget ablations, broad
fixed-snapshot repeats, complete threshold and guardrail provenance, and
immutable offline-to-online lineage. The available online evidence consists of
summarized relative changes rather than full randomization records, confidence
intervals, or candidate denominators, and the lower-cost model migration is
observational. The formal evaluation covers one application and two
recommendation scenarios; transfers to search, user growth, commercialization,
and other systems remain outside the evaluated evidence. Priority next steps
are protocol-matched repeats, fail-closed calibration, immutable lineage, and
matched component evaluations.

\section{Conclusion}
\label{sec:conclusion}

We presented \textbf{AutoLR}, an evidence-governed production harness that
automates the upstream stages of the recommendation research-to-review path,
from evidence retrieval and multi-expert proposal generation to direction
selection, repository-grounded implementation, training, offline evaluation,
and candidate packaging. LLM agents provide semantic reasoning and code
generation, while deterministic controllers retain authority over execution,
verification, guardrails, and persistent offline state. Online A/B admission
and full-traffic rollout remain human-gated in the deployment studied here.

A path-aware audit of the experiment logs recovered
\textbf{1,586 unique completed evaluations} across the single- and
two-column feed and the immersive-video feed. Nine production Launch
Review records reported positive online movements for engineer-selected
candidate packages. Routine AutoLR iterations were also migrated to a mixed
DeepSeek-V4-Pro/Flash stack at an observed LLM API cost of
\textbf{RMB 3--4 per iteration}, excluding model-training compute and internal
infrastructure. Together, these observations demonstrate the operational
viability of the integrated lifecycle, but do not isolate the causal
contribution of its individual components.

Long-running operation also revealed the \textbf{KEEP ratchet}: because each
KEEP changes the baseline for subsequent experiments, a non-reproducible
promotion can propagate into later comparisons and create apparent progress
that may not persist. The broader implication is that autonomous
experimentation must govern not only how broadly it searches, but also how
cautiously it commits. Durable autonomy therefore depends on structured
evidence and memory, deterministic verification, conservative state
transitions, and explicit human authority over production deployment.

\FloatBarrier
\balance
\bibliographystyle{ACM-Reference-Format}
\bibliography{autolr_references}

\end{document}